\pdfoutput=1

\documentclass[11pt]{article}

\usepackage[final]{acl}

\usepackage{times}
\usepackage{latexsym}

\usepackage[T1]{fontenc}

\usepackage[utf8]{inputenc}

\usepackage{microtype}

\usepackage{inconsolata}

\usepackage{graphicx}

\usepackage[T1]{fontenc}    
\usepackage{hyperref}       
\usepackage{url}            
\usepackage{booktabs}       
\usepackage{amsfonts}       
\usepackage{nicefrac}       
\usepackage{microtype}      
\usepackage{xcolor}         
\usepackage{natbib}         
\usepackage{amsmath}
\usepackage{subcaption}
\usepackage{multicol}
\usepackage{multirow}
\usepackage{cleveref}
\usepackage[first=0,last=9]{lcg}
\usepackage{color, colortbl}
\usepackage{graphicx}
\usepackage{amssymb}
\usepackage{xcolor,pifont}
\usepackage{listings}
\usepackage{color}

\definecolor{dkgreen}{rgb}{0,0.6,0}
\definecolor{gray}{rgb}{0.5,0.5,0.5}
\definecolor{mauve}{rgb}{0.58,0,0.82}

\newcommand{\greencheck}{{\color{green}\checkmark}}
\newcommand{\redx}{{\color{red}\ding{55}}}

\title{ULLME: A Unified Framework for Large Language Model Embeddings with Generation-Augmented Learning}

\author{{Hieu Man$^{1}$, Nghia Trung Ngo$^{1}$, Franck Dernoncourt$^2$, Thien Huu Nguyen$^1$,}\\
        $^1$Dept. of Computer Science, University of Oregon, OR, USA\\
        $^2$Adobe Research, USA\\
        \texttt{\{hieum,nghian,thienn\}@uoregon.edu},
        \texttt{franck.dernoncourt@adobe.com}
        }

\begin{document}
\maketitle
\begin{abstract}
Large Language Models (LLMs)\footnote{The definition of LLMs is vague. Here, we use ``LLMs'' to
refer to models with more than 1 billion parameters. Moreover, in the scope of this work, we focus on decoder-only LLMs.} excel in various natural language processing tasks, but leveraging them for dense passage embedding remains challenging. This is due to their causal attention mechanism and the misalignment between their pre-training objectives and the text ranking tasks. Despite some recent efforts to address these issues, existing frameworks for LLM-based text embeddings have been limited by their support for only a limited range of LLM architectures and fine-tuning strategies, limiting their practical application and versatility. In this work, we introduce the Unified framework for Large Language Model Embedding (ULLME), a flexible, plug-and-play implementation that enables bidirectional attention across various LLMs and supports a range of fine-tuning strategies. We also propose Generation-augmented Representation Learning (GRL), a novel fine-tuning method to boost LLMs for text embedding tasks. GRL enforces consistency between representation-based and generation-based relevance scores, leveraging LLMs’ powerful generative abilities for learning passage embeddings. To showcase our framework’s flexibility and effectiveness, we release three pre-trained models from ULLME with different backbone architectures, ranging from 1.5B to 8B parameters, all of which demonstrate strong performance on the Massive Text Embedding Benchmark. Our framework is publicly available at: \url{https://github.com/nlp-uoregon/ullme}. A demo video for ULLME can also be found at \url{https://rb.gy/ws1ile}.

\end{abstract}

\section{Introduction}

\begin{table}[t]
  \centering
   \resizebox{\columnwidth}{!}{
  \begin{tabular}{p{6cm}|c|ccc}

      \multicolumn{1}{c|}{\textbf{Framework}} & \multicolumn{1}{c|}{\textbf{\#Supported}} & \multicolumn{3}{c}{\textbf{Supported Fine-tuning Strategy}} \\ \cline{3-5}
      
       & \textbf{LLMs} & \textbf{SFT} & \textbf{DPO} & \textbf{Contrastive} \\ \hline
      SentenceTrasformers \cite{reimers-gurevych-2019-sentence} & >10 & \redx & \redx & \redx \\
      SGPT \cite{muennighoff2022sgpt} & 1 & \redx & \redx & \greencheck \\
      RepLLaMA \cite{ma2023finetuning} & 1 &  \redx & \redx & \greencheck \\
      Echo-Embedding \cite{springer2024repetition} & 2 & \redx & \redx & \redx \\
      GritLM \cite{muennighoff2024generative} & 2 & \greencheck & \redx & \greencheck \\
      LLM2Vec \cite{behnamghader2024llm2vec} & 3 & \redx & \redx & \greencheck \\ 
      NV-Emb \cite{lee2024nvembedimprovedtechniquestraining} & 1 & \redx & \redx & \greencheck \\ \hline
      ULLME (\textbf{our}) & >10 & \greencheck & \greencheck & \greencheck 
  \end{tabular}
   }
  \caption{Comparisions between ULLME and other LLM-Embedding frameworks. For ULLME,  the module combination enables many possible models and 10 is the number of models we have tested for usability.}
  \label{tab:compare}
\end{table}

For many years, the field of information retrieval has been dominated by a paradigm that relied heavily on pre-trained bidirectional encoders or encoder-decoders to obtain effective representation vectors for input texts (representation learning), e.g., BERT \cite{devlin2019bert} and T5 \cite{raffel2023exploring}. These architectures have played a pivotal role in advancing various language understanding tasks, including passage retrieval \cite{ni-etal-2022-sentence,qu-etal-2021-rocketqa, reimers-gurevych-2019-sentence}, \textit{inter alia}. However, recent research has witnessed a shift towards scaling representation learning methods to modern autoregressive language models \cite{muennighoff2022sgpt, muennighoff2024generative, behnamghader2024llm2vec}. Leveraging the ongoing advancements in large language models (LLMs) with various sizes and domains, this approach has the potential to transform research in information retrieval, significantly improving performance on related tasks.

However, directly applying pre-trained LLMs to dense retrieval still presents numerous challenges. These challenges primarily stem from two factors: the inherent limitations of LLMs' causal attention mechanism which restricts the models' attention to only preceding tokens \cite{muennighoff2022sgpt, springer2024repetition}, and the persistent misalignment between LLM pre-training objectives and text-ranking tasks \cite{ma2023finetuning, muennighoff2024generative, behnamghader2024llm2vec}. To address these issues, researchers have developed methods to enable bidirectional attention within LLMs by replacing the causal attention mask, which only allows attention to previous tokens, with an all-one mask that enables full contextual awareness. Furthermore, to better align the models with text retrieval tasks, researchers have employed fine-tuning strategies using retrieval-related data. However, as illustrated in Table\ref{tab:compare}, existing frameworks for LLM-based representation learning have been limited in their scope, supporting only a narrow range of LLM architectures and fine-tuning strategies. This limitation highlights the need for a flexible and comprehensive framework that can accommodate diverse combinations of LLM backbones and fine-tuning approaches to facilitate full explorations of possibilities in different areas.

In this paper, we present ULLME, a versatile and extensible platform designed to advance the use of LLMs for dense retrieval. ULLME addresses the critical limitations of existing frameworks by offering a comprehensive, plug-and-play solution that seamlessly enables bidirectional attention across a array of diverse LLM families, including LLaMa, Mistral, Phi, Qwen, among others. Our framework's flexibility also extends beyond model compatibility, supporting a wide spectrum of fine-tuning strategies for LLM-based representation learning. As such, ULLME provides an unified framework for various LLM backbones and fine-tuning methodologies, allowing developers to comprehensively explore the full potential of LLMs in diverse embedding tasks, free from the constraints of implementation-specific restrictions. 

In addition, existing frameworks for LLM-based text embeddings can be challenging for general users who are not familiar with training details like contrastive learning with large batch sizes and efficient fine-tuning. ULLME lowers these entry barriers by providing an efficient, user-friendly abstraction from those complexities, allowing users to focus on their data and tasks. For instance, ULLME’s training processes are integrated with advanced techniques like GradCache \cite{gao-etal-2021-scaling} and LoRa \cite{hu2022lora}, enabling efficient contrastive learning and tuning with larger batch sizes, and sparing users from complicated configuration and testing. ULLME also comes with user-friendly features that make it easy to evaluate various fine-tuned LLMs using the Massive Text Embedding Benchmark (MTEB) \cite{muennighoff-etal-2023-mteb}, a comprehensive evaluation suite with numerous tasks for text embeddings.

Building upon the ULLME framework, we further introduce Generation-augmented Representation Learning (GRL), a novel fine-tuning strategy that leverages LLMs' generative capabilities for enhanced passage embedding. GRL bridges traditional dense retrieval methods with LLMs' inherent generation strengths through two key mechanisms: (i) Joint Training: we simultaneously fine-tune LLMs on passage generation and contrastive learning tasks; (ii) Generation-Guided Representation Learning: we propose to directly leverage the passage's generation probabilities of LLMs to enhance representation learning. This is achieved by encouraging consistency between the passage-query cosine similarities (derived from learned embeddings) and the passages' generation probability of LLMs given the queries. GRL thus effectively aligns the understanding of LLMs for text relevance with respect to both the embedding and generation spaces, leading to more nuanced and richer embeddings from LLMs.


To showcase the versatility and effectiveness of ULLME, we release three pre-trained LLM-Embedding models with different backbone architectures, ranging from 1.5B to 8B parameters, which deliver highly competitive results on MTEB. Our findings also highlight the advantages of our new fine-tuning method, GRL, which significantly outperforms the strong baselines, underscoring the potential of our framework to advance research and development in LLM-based embeddings.

\section{Related Work}
Our work is situated within the field of Information Retrieval (IR), specifically focusing on frameworks that leverage Large Language Models (LLMs) for Dense Retrieval.

\textbf{LLMs for Dense Retrieval.}  Recent advancements in this area have primarily addressed two key challenges: (i): Overcoming LLMs' Causal Attention Limitations by developing methods to enable bidirectional attention within LLMs \cite{muennighoff2022sgpt, muennighoff2024generative, behnamghader2024llm2vec, lee2024nvembedimprovedtechniquestraining}, allowing models to consider both past and future context when computing embeddings, and (ii): Aligning LLM Pre-training with Text Ranking by fine-tuning LLMs via contrastive learning \cite{ma2023finetuning, wang2024improving, 
lee2024nvembedimprovedtechniquestraining}. This process can also be augmented with additional objectives such as supervised fine-tuning (SFT) \cite{muennighoff2024generative} or mask-filling tasks \cite{ behnamghader2024llm2vec}. An alternative approach proposed by \citet{springer2024repetition} involves a prompting method where the input sequence is duplicated, enabling each token to attend to future tokens and mitigating the contextualization issues inherent in causal attention. While these methods have shown promise, they generally do not explicitly enforce consistency between the model's understanding of relevance in both the embedding and generation spaces. This limitation restricts their ability to fully leverage the remarkable generative capabilities of LLMs for dense retrieval tasks. Our work, GRL, builds upon these foundations while addressing their limitations, introducing novel techniques to harmonize embedding-based and generation-based relevance scoring within a unified framework.

\textbf{Frameworks of LLMs for Dense Retrieval.} 
Existing frameworks for LLMs in Dense Retrieval have been constrained by their limited support for LLM architectures and fine-tuning strategies. As shown in Table\ref{tab:compare}, SentenceTransformers\cite{reimers-gurevych-2019-sentence} supports various types of LLMs but is primarily designed for inference without allowing fine-tuning, limiting its applicability in advancing state-of-the-art dense retrieval methods. Some recent works \cite{muennighoff2022sgpt, ma2023finetuning, lee2024nvembedimprovedtechniquestraining}, such as \textbf{Echo} \cite{wang2024improving}, \textbf{GritLM} \cite{muennighoff2024generative}, \textbf{LLM2Vec} \cite{behnamghader2024llm2vec}, and the models in the Hugging Face's MTEB leaderboard\footnote{\url{https://huggingface.co/spaces/mteb/leaderboard}}, have introduced implementations for LLM-based text embeddings. However, these approaches are often tailored to specific model architectures and training methods with hard-coded implementations, thus restricting their adaptability and use across different LLM architectures and fine-tuning strategies to meet diverse development and application demands. In contrast, our framework ULLME addresses these limitations by offering a flexible and extensible platform. ULLME can accommodate a diverse range of LLM backbones and supports various training approaches, making it highly versatile and broadly applicable. 

\section{ULLME - Unified framework for Large Language Model Embedding}
We present an overview of our ULLME framework in Section \ref{sec:Overview} while Section \ref{sec:features} details the key technical methods.

\begin{lstlisting}[float=t, captionpos=b, caption={Extending bidirectional attention for LLMs via ULLME.}]
from ullme.models import ULLME

model = ULLME(
    model_name_or_path="mistralai/Mistral-7B-v0.1",
    model_backbone_type="mistral",
    lora_name="ullme-mistral",
    loar_r=16,
    lora_alpha=32,
    )
input_sentence = "This a example sentence."
model_inputs = model.tokenizer(
    [input_sentence], 
    return_tensors='pt'
    )
model_output = model(
    input_ids=model_inputs['input_ids'],
    attention_mask=model_inputs['attention_mask'],
    is_generate=False
    )
>> {'rep': (1, hidden_dim)}
\end{lstlisting}

\subsection{Overview}
\label{sec:Overview}
ULLME addresses the limitations of existing LLM-based dense retrieval frameworks by offering a flexible and comprehensive solution. The framework operates in three main stages. First, it enables bidirectional attention within LLMs by replacing the causal attention mask with a bidirectional one. This crucial modification extends the models' ability to consider both past and future context when generating embeddings, significantly enhancing its capacity for dense retrieval tasks. The transformed model is then returned as a PyTorch object, providing users with the flexibility to integrate it into various frameworks or pipelines. We will elaborate on this process in Section \ref{sec:bidirectional_attn}. Second, ULLME supports a diverse array of fine-tuning strategies, including Contrastive Learning, Supervised Fine-tuning (SFT), Direct Preference Optimization (DPO), and our novel Generation-augmented Representation Learning (GRL). This versatility allows for tailored optimization across a wide spectrum of retrieval tasks and domains, as detailed in Section \ref{sec:finetune_stratergies}. Finally, the framework streamlines the evaluation process by incorporating direct support for model validation using the Massive Text Embedding Benchmark (MTEB) library (Section \ref{sec:eval_process}). This integration facilitates comprehensive assessment across numerous retrieval and embedding tasks. By seamlessly combining these elements, ULLME provides an extensive toolkit for leveraging LLMs in diverse dense retrieval tasks, encompassing everything from initial model adaptation to fine-tuning and evaluation. Our comprehensive approach aims to accelerate research and development for of LLM-based dense retrieval, offering researchers and practitioners a comprehensive platform for innovation and advancement.

\subsection{Key Features}
\label{sec:features}
\subsubsection{Enabling Bidirectional Attention}
\label{sec:bidirectional_attn}

To enable bidirectional attention in LLMs, ULLME requires only minimal code modifications, as illustrated in Listing 1. The framework's user-friendly design allows for easy initialization with various LLM backbones by simply specifying the \texttt{``model\_name\_or\_path''} and \texttt{``model\_backbone\_type''} parameters. ULLME seamlessly integrates with Hugging Face Transformers, loading pre-trained LLMs directly from their repository. Additionally, our framework supports parameter-efficient fine-tuning through Low-Rank Adaptation (LoRA) \cite{hu2022lora}, offering flexibility in model adaptation. Once initialized, the model can be used to compute sequence representations. The \texttt{``is\_generate''} parameter plays a crucial role in controlling the attention mechanism: when set to \texttt{``False''}, the model employs bidirectional attention, optimizing it for dense retrieval tasks, while \texttt{``True''} reverts the model to causal attention, mimicking the standard Hugging Face Transformer model output. This dual functionality allows ULLME to serve both as an advanced specialized embedding model and as a language model when needed, providing developers with a flexible tool that can conveniently transition between bidirectional and causal attention modes. ULLME provides various methods for extracting text embeddings from LLMs, such as using representations from the first token, last token, mean, or weighted mean pooling. However, it defaults to averaging the representation vectors from the final layers (mean) for better performance on our datasets.


\subsubsection{Fine-tuning Strategies}
\label{sec:finetune_stratergies}
\begin{lstlisting}[float=t, captionpos=b, caption={Finetuning LLMs for text embedings via ULLME.}]
from ullme.trainer import GradCacheTrainer

trainer = GradCacheTrainer(
    con_loss_type='NTXentLoss',
    gen_loss_type='dpo', # 'sft'
    use_kl_loss=True
)
trainer.fit_epoch(
    model=model,
    train_loader=train_dataloader,
)
\end{lstlisting}

Our ULLME framework supports multiple fine-tuning strategies, as illustrated in Listing 2.

\textbf{Contrastive Learning.} ULLME's Contrastive Learning objective utilizes in-batch negatives \cite{10.5555/3524938.3525087, gao-etal-2021-simcse}. The contrastive loss is formally defined as: $\mathcal{L}_{CL} = \\
    -\log\frac{\exp{(s_{rt}(q, p^+))}}{\exp{(s_{rt}(q, p^+))} + \sum_{p^-\in B} \exp{(s_{rt}(q, p^-))}} \notag$.
    
Here, $B$ represents a mini-batch, $q$ is the input query, $p^+$ denotes the positive (relevant) passage, and $p^-$ represents negative (non-relevant) passages sampled from the current training mini-batch. The function $s_{rt}(q, p)$ computes the relevance score between a query and a passage using cosine similarity of the induced representations for $q$ and $p$. To enhance the effectiveness of Contrastive Learning, especially under limited GPU memory constraints, ULLME incorporates advanced techniques such as GradCache \cite{gao-etal-2021-scaling} and cross-device contrastive loss computation. These optimizations allow for efficient training with larger batch sizes and more diverse negative samples, which are crucial for learning high-quality representations.

\textbf{Supervised Fine-tuning (SFT).} In addition to contrastive learning, ULLME supports SFT, a strategy that enhances LLMs' ability to generate high-quality passages in response to queries. ULLME implements SFT using a next-word prediction objective: $\mathcal{L}_{SFT} = -\frac{1}{N}\sum_{i=1}^N \log\pi_\theta(w_i|w_{< i}, q) \notag $. Here, $N$ is the length of the positive passage $p^+$, $w_i$ is the $i$-th token in $p^+$, and $\pi_\theta(w|x)$ is the conditional likelihood of $w$ given $x$, computed by the LLM $\theta$. Importantly, during SFT loss computation, ULLME reverts to using causal attention, mirroring standard LLM behavior.

\textbf{Direct Preference Optimization (DPO).} 
ULLME incorporates Direct Preference Optimization (DPO) \cite{NEURIPS2023_a85b405e} as an advanced fine-tuning strategy, offering an alternative to traditional Supervised Fine-tuning (SFT). DPO has demonstrated superior effectiveness in LLM fine-tuning. Moreover, the DPO approach inherently accounts for both preferred and rejected outputs, making it intuitively more suitable for aligning models with text-ranking objectives compared to SFT. In ULLME's implementation, the ground-truth relevant passage $p^+$ for a query $q$ is treated as the preferred output, while negative and irrelevant passages $p^-$ are considered dispreferred. The DPO loss function is designed to encourage the model to assign higher generation probabilities to $p^+$ compared to any $p^-$: $\mathcal{L}_{DPO}  = \\ 
- \log\sigma\left(\beta\log\frac{\pi_\theta(p^+|q)}{\pi_{ref}(p^+|q)} - \beta\log\frac{\pi_\theta(p^-|q)}{\pi_{ref}(p^-|q)}\right) \notag$.
In this formulation, $\sigma$ represents the sigmoid function, $\beta$ is a scaling factor, and $\pi_{ref}(p|q)$ denotes the conditional likelihood computed by the original pre-trained LLM (the reference model). 

In addition to the standard DPO formulation, ULLME includes implementations of advanced variants such as Kahneman-Tversky Optimization (KTO) \cite{ethayarajh2024kto} and Contrastive Preference Optimization (CPO) \cite{xu2024contrastive}. The modular architecture of ULLME facilitates the seamless integration of new preference optimization techniques as they emerge, ensuring that the framework remains at the forefront of LLM fine-tuning advancements. Finally, to maintain consistency with the model's pre-training paradigm, ULLME employs causal attention when computing the DPO loss, similar to the approach used in SFT.  

\textbf{Generation-augmented Representation Learning (GRL).} ULLME further introduces a novel fine-tuning strategy GRL that explicitly aligns the LLMs' understanding of passage-query text relevance in embedding and generation spaces to boost representation learning. As such, GRL first computes a generation-based relevance score $s_{gen}(q, p)$ utilizing the conditional generation likelihood of a passage candidate $p$ given input query $q$ from LLMs: $s_{gen}(q, p) = \frac{1}{t}\sum_{i=1}^t \log\pi_\theta(w_i|w_{< i}, q) \notag $,
where $t$ is the length of $p$ and $w_i$ is the $i$-th token in $p$. 

Next, we seek to recognize the consistency of the query-passage relevance scores obtained from the representations (i.e., $s_{rt}(q,p)$) and the generation likelihood (i.e., $s_{gen}(q,p)$). Particularly, let $U$ be the set of $m$ candidate passages for $q$. For each candidate passage $p_i \in U$, we compute $s_{rt}(q, p_i)$ and $s_{gen}(q, p_i)$, then normalize these scores to obtain the representation and generation relevance distributions over $U$: $P_{rt}(q, p_i) = \frac{\exp(s_{rt}(q, p_i))}{\sum_{p' \in U} \exp(s_{rt}(q, p'))}$ and $P_{gen}(q, p_i) = \frac{\exp(s_{gen}(q, p_i))}{\sum_{p' \in U} \exp(s_{gen}(q, p'))}$.

Afterward, we minimize the KL divergence between their distributions: $\mathcal{L}_{KL} = \sum_{p \in U} P_{rt}(q, p) \log \frac{P_{rt}(q, p)}{P_{gen}(q, p)} \notag$, serving as a training signal to enrich representation learning for LLMs.

Finally, the overall training loss for GRL combines the contrastive loss $\mathcal{L}_{CL}$, the direct preference optimization loss $\mathcal{L}_{DPO}$, and the KL-divergence loss $\mathcal{L}_{KL}$: $\mathcal{L}_{GRL} = \lambda_{CL} \mathcal{L}_{CL} + \lambda_{DPO} \mathcal{L}_{DPO} + \lambda_{KL} \mathcal{L}_{KL} \notag$,
where $\lambda_{CL}$, $\lambda_{DPO}$, and $\lambda_{KL}$ are weighting hyperparameters.

\subsection{Evaluation Process}
\label{sec:eval_process}
\begin{lstlisting}[float=t, captionpos=b, caption={Evaluation on MTEB dataset via ULLME.}]
from ullme.models import WrappedULLME
from ullme.eval import eval_mteb_dataset

model = WrappedULLME(
    model_name_or_path="mistralai/Mistral-7B-v0.1",
    model_backbone_type="mistral",
    lora_name="ullme-mistral",
    loar_r=16,
    lora_alpha=32,
    model_checkpoint="path/to/your/checkpoint"
    )
eval_result = eval_mteb_dataset(
    model=model,
    dataset_name='MSMARCO',
    langs=['eng'],
    )
>> {'eng': 35.8}
\end{lstlisting}

ULLME streamlines the evaluation process by integrating direct support for evaluating LLM-based text embedding models over MTEB\footnote{\url{https://github.com/embeddings-benchmark/mteb}}, a widely-used Massive Text Embedding Benchmark with diverse tasks and datasets. This integration facilitates comprehensive model development with different methods and extensive assessment across numerous retrieval and embedding tasks in a single framework. ULLME wraps a fine-tuned model into a \texttt{``WrappedULLME''} instance, ensuring compatibility with MTEB's requirements for direct evaluation. In addition to supporting ULLME's fine-tuned models, our evaluation function is designed to perform seamlessly with most LLM models available in the Hugging Face ecosystem, including the latest LLM-Embedding models in the MTEB leaderboard. Users can easily specify the desired model through the \texttt{``model\_name\_or\_path''} parameter, enabling effortless evaluation of various LLMs without the need for extensive configuration. ULLME allows users to select specific datasets and language subsets for evaluation. The evaluation results are reported using MTEB's predefined main scores of the corresponding dataset, ensuring standardized and comparable metrics across different models, as demonstrated in Listing 3.

\section{Experiments}

Our ULLME framework supports various LLM architectures and fine-tuning strategies for text embeddings with convenient interface. To highlight the framework's flexibility, we demonstrate the operations of ULLME with three different base LLMs ranging from 1.5B to 8B parameters: Phi-1.5B \cite{textbooks2}, Mistral-7B-Instruct-v0.2 \cite{jiang2023mistral}, and Meta-LLama3-8B-Instruct \cite{llama3modelcard}. For each LLM, we evaluate ULLME's performance for different combinations of attention and fine-tuning approaches, including: \textbf{Base}: Original causal model, \textbf{Causal + CL}: Causal model fine-tuned with Contrastive Learning, \textbf{Bi + CL}: Bidirectional-enabled model fine-tuned with Contrastive Learning, and \textbf{Bi + CL + SFT}:  Bidirectional-enabled model fine-tuned with Contrastive Learning and SFT. In addition, we report the performance of our Generation-augmented Representation Learning (GRL) method for fine-tuning LLMs in ULLME, featuring the full model GRL and \textbf{$\text{GRL}_{SFT}$}, a variant of GRL that replaces DPO with SFT for tuning. Finally, we compare the performance of ULLME's models with recent state-of-the-art methods for LLM-based text embeddings, including \textbf{Echo} \cite{wang2024improving} and \textbf{LLM2Vec} \cite{behnamghader2024llm2vec}.

\textbf{Settings.} Following prior work \cite{qu-etal-2021-rocketqa,ren-etal-2021-rocketqav2,ma2023finetuning}, we use a curated subset of the MSMARCO dataset \cite{bajaj2018ms} for model training. MTEB datasets are employed for evaluation. To train the models, we utilize LoRA \cite{hu2022lora} with rank of 16, and enable various optimization techniques, i.e., GradCache, gradient checkpointing, mixed precision training, and FSDP \cite{zhao2023pytorch}, to minimize GPU memory requirements. We utilize the AdamW optimizer \cite{Loshchilov2017DecoupledWD} with a learning rate of 2e-4 and a batch size of 512 for one epoch on MSMARCO. The weights for the GRL loss components include $\lambda_{CL}=\lambda_{KL}=1$ and $\lambda_{DPO}=0.5$. A reproducibility checklist is included in Appendix \ref{app:repo}.

\begin{table}[t]
  \centering
  \small
  \resizebox{\columnwidth}{!}{
  \begin{tabular}{lccc}
  \toprule
  \textbf{ } & \texttt{Phi 1.5} & \texttt{Mistral-2-7B} & \texttt{LlaMa-3-8B} \\
  \midrule
  Echo* & 36.00 & 50.26 & 51.11\\ %
  LLM2Vec$^*$ & 54.47 & 57.47 & 58.04\\ \hline
  Base & 31.15 & 42.31 & 42.33 \\
  Causal + CL & 51.83 & 54.03 & 54.68\\
  Bi + CL & 52.70 & 55.41 & 55.86\\
  Bi + CL + SFT & 53.88 & 57.01 & 56.83 \\ \hline
  $\text{GRL}_{SFT}$ & 55.01 & 58.37  & 57.50 \\
  GRL (ours) & \bf{55.76} & \bf{59.50} & \bf{59.27}\\
  \bottomrule
  \end{tabular}
  }
  \caption{Model performances on MTEB datasets using MSMARCO for training data. The numbers are averaged over 56 datasets of MTEB, covering diverse tasks such as Retrieval, Reranking, Clustering, Pair Classification, Classification, Semantic Textual Similarity, and Summarization. The best results are in bold and $^*$ indicates our implementation/reproduced results using the same training data. Detailed performance for all datasets in MTEB is reported in Table \ref{tab:mteb-results}.}
  \label{tab:mteb-msmarco}
\end{table}

\textbf{Results.} Table \ref{tab:mteb-msmarco} showcases the performance of various models on the MTEB datasets. Compared to previous methods Echo and LLM2Vec, it is clear that our ULLME framework can be used to train diverse and competitive LLM-based embedding models for different base LLMs and tasks in MTEB. Among various architectures in ULLME, we observe that the combination of contrastive learning and SFT leads to better performance than the individual techniques, demonstrating their complementary benefits for LLM-based embeddings. Notably, our proposed Generation-augmented Representation Learning (GRL) method in ULLME consistently outperforms the best baseline, LLM2Vec, across different base models ranging from 1.5B to 8B parameters. This highlights the effectiveness of using generation probabilities to guide representation learning in GRL. Finally, we note that the inference time of the fine-tuned models with ULLME is comparable to the original LLMs, processing 16K, 12K, and 12.8K tokens per second for Phi-1.5B, Mistral-7B-Instruct-v0.2, and Meta-LLama3-8B-Instruct, respectively.

\section{Conclusion}

We introduce ULLME (Unified framework for Large Language Model Embedding), a comprehensive and flexible toolkit for leveraging LLMs for text embeddings and dense retrieval tasks. Our work addresses critical limitations in existing frameworks for LLM embeddings by providing support for various LLM architectures, fine-tuning strategies, and benchmark evaluation within a single, user-friendly framework. Our experimental results demonstrate the effectiveness of ULLME, particularly the GRL strategy, in improving dense retrieval performance across various LLM scales and tasks. Our potential future directions include exploration of better techniques to leverage the generative and discriminative capabilities of LLMs, and extension of the framework to support emerging LLM architectures and training paradigms. 

\bibliography{custom}
\newpage
\clearpage
\appendix
\section{Reproducibility Checklist}
\label{app:repo}

\begin{itemize}
\item \textbf{Source code with specification of all dependencies, including external libraries}: Our source code, along with a README file detailing all dependencies and external libraries, is publicly available here: \url{https://github.com/nlp-uoregon/ullme}.

\item \textbf{Description of computing infrastructure used}: Experiments were conducted on a computing infrastructure comprising 4 NVIDIA A100 GPUs with 80GB of memory each. We utilized PyTorch 2.1.1 and the Hugging Face Transformers library (version 4.35.0) for model implementation and training.

\item \textbf{Average runtime}: The fine-tuning process with GRL for the 1.5B models took approximately 1.5 days, while the process for the larger 7B and 8B models required 5 and 6 days, respectively. The reported results represent the average performance with 3 random seeds.

\item \textbf{Explanation of evaluation metrics used, with links to code}: We employed a comprehensive set of evaluation metrics, including nDCG@10, MAP, V-Means, AP and Accuracy for retrieval, reranking, clustering, classification tasks. For STS and summarization tasks, we use Spear to measure the performance. The evaluation was performed by our framework ULLME, which follows the standards for the MTEB datasets \cite{muennighoff-etal-2023-mteb}. Finally, we used the same instructions as described in \cite{wang2024improving} for the tasks in MTEB during our evaluation.

\item \textbf{Hyperparameter configurations}: All models were trained with the AdamW optimizer, a learning rate of $2e^{-4}$, and a batch size of 512. The training was conducted for 1 epoch on the MSMARCO dataset, corresponding to 545 training steps. The number of hard negative passages per example was set to 8. The scaling factor $\beta$ in the DPO loss was set to 0.1. The weights for the overall loss components in GRL were $\lambda_{CL}=\lambda_{KL}=1$ and $\lambda_{DPO}=0.5$. For all models, we employed LoRA with $r=16$, $\alpha=32$, and a dropout probability of 0.2.
\end{itemize}

\section{Detailed Performance on MTEB}

We present the full performance of the three ULLME-released models -- Phi-1.5 \cite{textbooks2}, Mistral-2-7B-instruct \cite{jiang2023mistral}, and LLaMa-3-B-instruct \cite{llama3modelcard} -- across the MTEB datasets in Table \ref{tab:mteb-results}.

\begin{table*}[t]
    \centering
    \small
    \begin{tabular}{l|ccc}
    \toprule
    \textbf{Task} & \texttt{Phi 1.5} & \texttt{Mistral-2-7B} & \texttt{LlaMa-3-8B} \\
    \midrule
    AmazonCounterfactualClassification & 67.79 & 75.28 & 73.69 \\
    AmazonPolarityClassification & 72.03 & 77.40 & 78.51 \\
    AmazonReviewsClassification & 35.58 & 39.78 & 38.31 \\
    Banking77Classification & 84.24 & 84.57 & 84.76 \\
    EmotionClassification & 45.83 & 45.02 & 49.48 \\
    ImdbClassification & 66.73 & 72.47 & 74.97 \\
    MassiveIntentClassification & 70.43 & 73.41 & 73.1 \\
    MassiveScenarioClassification & 76.75 & 78.28 & 78.59 \\
    MTOPDomainClassification & 92.58 & 94.72 & 94.70 \\
    MTOPIntentClassification & 69.63 & 77.05 & 73.49 \\
    ToxicConversationsClassification & 66.26 & 60.62 & 64.21 \\
    TweetSentimentExtractionClassification & 55.92 & 55.99 & 56.63 \\
    ArxivClusteringP2P & 42.29 & 46.97 & 46.46 \\
    ArxivClusteringS2S & 31.65 & 39.92 & 37.91 \\
    BiorxivClusteringP2P & 36.25 & 38.18 & 38.35 \\
    BiorxivClusteringS2S & 30.46 & 31.48 & 30.32 \\
    MedrxivClusteringP2P & 31.82 & 32.32 & 32.19 \\
    MedrxivClusteringS2S & 30.18 & 26.95 & 26.01 \\
    RedditClustering & 49.31 & 41.45 & 41.96 \\
    RedditClusteringP2P & 55.85 & 62.26 & 61.64 \\
    StackExchangeClustering & 60.6 & 62.44 & 61.06 \\
    StackExchangeClusteringP2P & 31.79 & 32.99 & 33.77 \\
    TwentyNewsgroupsClustering & 42.95 & 38.52 & 41.32 \\
    SprintDuplicateQuestions & 92.78 & 92.2 & 94.73 \\
    TwitterSemEval2015 & 59.19 & 67.35 & 69.0 \\
    TwitterURLCorpus & 85.06 & 86.81 & 85.61 \\
    AskUbuntuDupQuestions & 59.23 & 63.62 & 63.43 \\
    MindSmallReranking & 31.70 & 32.30 & 31.66 \\
    SciDocsRR & 79.29 & 83.47 & 81.42 \\
    StackOverflowDupQuestions & 48.61 & 52.56 & 52.38 \\
    ArguAna & 55.06 & 45.93 & 46.78 \\
    ClimateFEVER & 22.28 & 28.10 & 22.22 \\
    CQADupstackTexRetrieval & 22.39 & 25.84 & 28.30 \\
    DBPedia & 30.45 & 46.55 & 46.36 \\
    FEVER & 58.11 & 79.39 & 61.52 \\
    FiQA2018 & 32.25 & 42.97 & 42.28 \\
    HotpotQA & 48.44 & 64.04 & 67.41 \\
    MSMARCO & 28.65 & 34.22 & 35.65 \\
    NFCorpus & 34.54 & 39.37 & 39.37 \\
    NQ & 38.37 & 60.73 & 61.36 \\
    QuoraRetrieval & 86.49 & 88.33 & 87.75 \\
    SCIDOCS & 16.46 & 21.00 & 21.13 \\
    SciFact & 63.41 & 72.86 & 72.38 \\
    Touche2020 & 16.56 & 30.52 & 27.13 \\
    TRECCOVID & 54.21 & 84.74 & 83.56 \\
    BIOSSES & 85.35 & 78.64 & 83.74 \\
    SICK-R & 70.49 & 70.31 & 69.11 \\
    STS12 & 71.83 & 67.25 & 69.95 \\
    STS13 & 80.05 & 82.35 & 79.58 \\
    STS14 & 74.19 & 75.04 & 73.67 \\
    STS15 & 83.0 & 82.69 & 83.47 \\
    STS16 & 79.77 & 81.15 & 81.58 \\
    STS17 & 88.49 & 86.38 & 86.3 \\
    STS22 & 67.77 & 68.54 & 67.35 \\
    STSBenchmark & 80.81 & 78.21 & 80.25 \\
    SummEval & 30.61 & 30.56 & 31.10 \\
    \midrule
    Average & 55.76 & 59.50 & 59.27 \\
    \bottomrule
    \end{tabular}
    \caption{Performance of ULLME's released models on full MTEB benchmark using MSMARCO as training data.}\label{tab:mteb-results}
\end{table*}

\end{document}